\pdfoutput=1

\documentclass[11pt]{article}

\usepackage{acl}

\usepackage{times}
\usepackage{latexsym}

\usepackage[T1]{fontenc}

\usepackage[utf8]{inputenc}

\usepackage{microtype}

\usepackage{inconsolata}

\usepackage{graphicx}
\usepackage{tikz}
\usepackage{xcolor}
\usepackage{colortbl}
\usepackage{epsdice}
\usetikzlibrary{decorations.markings}
\newcommand{\ok}{\cellcolor{green!10}{\tikz[scale=0.2] {
    \fill[green!10] (0,-1) rectangle (1.2,1);
    \draw[line width=0.7mm, green!50!black] (0,0) -- (0.3,-0.5) -- (1.1, 0.5); 
}}}\newcommand{\hm}{\cellcolor{yellow}{\tikz[scale=0.2] {
    \fill[yellow] (0,-1) rectangle (1.2,1);
    \draw[line width=0.7mm, yellow!50!black] (0,0) -- (0.3,-0.5) -- (1.1, 0.5); 
}}}
\newcommand{\no}{\cellcolor{red!10}{\tikz[scale=0.2] {
  \draw[line width=0.7mm, red!50!black] (0,0) -- (1,1);
  \draw[line width=0.7mm, red!50!black] (0,1) -- (1,0);
}}}

\title{Large Language Models and Mathematical Reasoning Failures}
\author{Johan Boye \\
  KTH Royal Institute of Technology\\
  Stockholm, Sweden \\
  \texttt{jboye@kth.se} \\\And
  Birger Moëll \\
  KTH Royal Institute of Technology\\
  Stockholm, Sweden \\
  \texttt{bmoell@kth.se} \\}

\begin{document}

\maketitle

\begin{abstract}
This paper investigates the mathematical reasoning capabilities of large language models (LLMs) using 50 newly constructed high-school-level word problems. Unlike prior studies focusing solely on answer correctness, we rigorously analyze both final answers and solution steps to identify reasoning failures. Evaluating eight state-of-the-art models—including Mixtral, Llama, Gemini, GPT-4o, and OpenAI’s o1 variants—we find that while newer models (e.g., o3-mini, deepseek-r1) achieve higher accuracy, all models exhibit errors in spatial reasoning, strategic planning, and arithmetic, sometimes producing correct answers via flawed logic. Common failure modes include unwarranted assumptions, over-reliance on numerical patterns, and inability to translate physical intuition into mathematical steps. Manual scrutiny reveals that models struggle with problems requiring multi-step deduction or real-world knowledge, despite possessing broad mathematical knowledge. Our results underscore the importance of evaluating reasoning processes, not just answers, and caution against overestimating LLMs’ problem-solving proficiency. The study highlights persistent gaps in LLMs’ generalization abilities, emphasizing the need for targeted improvements in structured reasoning and constraint handling.
\end{abstract}

\section{Introduction}
How good are large language models (LLMs) at mathematical reasoning? This question has been addressed by several authors, who have constructed data sets in order to evaluate the mathematical capabilities of LLMs, e.g.\  \citep{hendrycks2020measuring, hendrycks2021measuring, cobbe2021training, chernyshev2024u, li2024gsm}. In most of these studies, only the final answer produced by the LLM on a given problem was checked for correctness  -- the questions were either multiple-choice, or the answer consisted of a single number, both cases facilitating automatic evaluation. However, as it is possible to arrive at a correct answer by means of shallow heuristics rather than a watertight argument, it is important to also study the full solution provided by the model, much in the same way a teacher would assess a student exam. Of course, this method requires manual scrutiny and is therefore more time-consuming, but we argue that it is indispensable for to get a proper picture of the mathematical prowess of LLMs.

In this paper, we present a small dataset\footnote{Dataset, code, and results are available on \url{https://github.com/jboye12/llm-probs}} of 50 newly constructed mathematical problems intended for LLM evaluation, and use it to evaluate X models: mixtral8x7b \citep{jiang2024mixtralexperts}, llama3.3-70B-versatile \citep{touvron2023llamaopenefficientfoundation}, Gemini-2.0-pro-exp \citep{gemini-2.0}, GPT4o \cite{openai2024gpt4technicalreport}, o1-preview, o1, and o3-mini \cite{openaio1}. Our problems are all formulated in natural language (``word problems'') and require no more than high-school level mathematical knowledge: basic principles of counting and divisibility, some algebra, arithmetic, probability and geometry, and some real-world knowledge, e.g.\  that it is impossible to walk on water, how many minutes there are in an hour, how the dots are placed on dice (e.g.\ the \epsdice{1} is opposite the \epsdice{6}), and so on. We purposely excluded complicated sums or integrals written in pure mathematical notation, since there are already computer algebra systems like Mathematica that can solve large classes of such problems in a precise way. Our goal was rather to focus on natural language word problems.

Such mathematical word problems provide an excellent testbed for evaluating the reasoning capabilities of LLMs. Early LLMs were not explicitly trained to perform reasoning but rather to do next-token prediction, possibly with additional training based on techniques like RLHF \citep{ouyang2022training}. Still, these models seemed capable of performing non-trivial reasoning in many instances, in particular when prompted with a ``Chain-of-Thought'' prompt like {\em Let's think step by step} \citep{wei2022chain}. However, it is not clear how much of these apparent reasoning capabilities can be attributed to {\em memorization\/} of the training material combined with shallow heuristics, as opposed to having learned actual general principles of reasoning by generalizing from the training examples. \citet{prabhakar2024decipheringfactorsinfluencingefficacy} conclude that it is a combination of probabilistic, noisy reasoning and memorization of the training material, and  the more reasoning steps are required to get to the solution, the more likely it is that memorization will interfere with the reasoning process, leading to the wrong answer.

Starting in the fall of 2024, several models were released that more explicitly combined next-token prediction with reasoning. In the announcement of their "o1" models, OpenAI write: {\em In a qualifying exam for the International Mathematics Olympiad (IMO), GPT-4o correctly solved only 13\% of problems, while the reasoning model scored 83\%} \citep{openaio1}. This claim somewhat mirrored by our results, with the o1 model achieving 37/50 on our problem set. We still found it somewhat surprising that o1 was not better still, considering that our problems are far easier than the typical IMO problems. In the paper, we make a systematic study of the reasoning failures exhibited by various models, and try to analyse their root causes.

\section{Related work} 
A number of researchers have created datasets to evaluate the mathematical abilities of LLMs. {\bf MATH} \citep{hendrycks2021measuring} contains a large collection of mathematical problems from different domains, with 7 different levels of difficulty. The answer is always a number. Also the MMLU \citep{hendrycks2020measuring} contains a mathematics section consisting of multiple-choice questions.  

{\bf GSM8K} \citep{cobbe2021training} contains word problems on a grade-school level solvable by simple arithmetic. The answer is always a number. {\bf GSM-Plus} \citep{li2024gsm} and  {\bf GSM-symbolic} \citep{mirzadeh2024gsm} are both extensions of GSM8k with adversarial examples. In the latter case, the authors showed that it was possible to confuse the models by adding irrelevant numerical information to the problem formulation. In some cases, the models worked this irrelevant information into the solutions, leading to incorrect answers. 

{\bf U-MATH} \citep{chernyshev2024u} contains university-level problems given in mathematical notation and with figures (i.e.\ the input is multimodal). 

All these datasets are quite large, containing thousands of similar problems, and they are amenable to automatic assessment.

\section{Method}
We constructed a set of 50 problems. Four problems were taken from a Swedish book of mathematical puzzles \citep{vaderlind1996}, the rest were invented by the authors. The selection/design criterion for the problems was that they should be solvable with only high-school mathematics, although the questions themselves might be of a different nature than those posed to high-school students (depending on country). Some problems had a specific numerical answer, some asked for a statement of the type "It is possible/impossible to do X?", and some asked for a concrete method, algorithm, or strategy to obtain some particular goal. All problems are listed in the appendix.

Each question was posed once to each model through their respective APIs\footnote{The {\bf mixtral} and {\bf llama} models were hosted at Groq ({\tt https://groq.com}) and called though the Groq API.} This led to 400 answers from the models, which were assessed manually by the first author (who is also an experienced teacher), checking both the answer and the solution for correctness. If the solution was incorrect, we also wrote a brief note describing the nature of the problem. 

\section{Results}

\begin{table}[h]
    \begin{center}
        \begin{tabular}{|l|c|c|c|}
        \hline
            {\bf Model} & {\bf Correct} & {\bf Ans} & {\bf Sol}  \\
            \hline
            mixtral-8x7B & 0 & 4 & 0 \\
            llama-3.3-70B & 10 & 1 & 0 \\
            gemini-2.0-pro-exp & 23 & 3 & 1 \\
            gpt-4o & 14 & 3 & 2 \\
            o1-preview & 30 & 2 & 2 \\
            o1 & 37 & 2 & 1 \\
            o3-mini & 40 & 2 & 0 \\
            deepseek-r1 & 36 & 4 & 0 \\
            \hline
        \end{tabular}
    \end{center}
    \caption{The number of problems correctly solved and answered (out of 50). {\bf Ans} = correct answer but wrong solution. {\bf Sol} = correct solution but wrong final answer.}
    \label{fig:results}
\end{table}

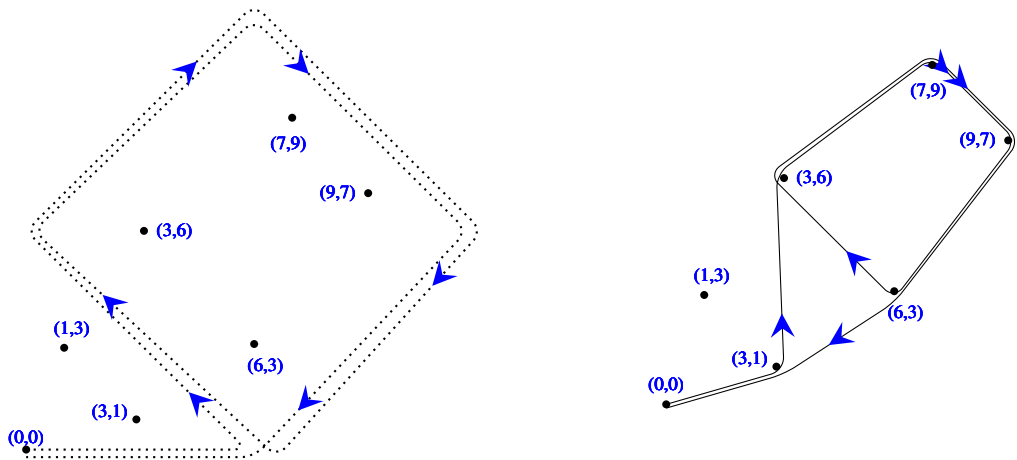
\begin{figure*}[t]
    \begin{tabular}{cc}
\begin{minipage}{0.5\textwidth}
\begin{scriptsize}
\begin{tikzpicture}[scale=0.5,
  decoration={
    markings,
    mark=between positions 0.1 and 1.0 step 0.16 with {\arrow[blue,line width=1mm]{stealth}}
  },
  rounded corners=5pt 
]

\coordinate (A) at (0,0.2);
\coordinate (B) at (5.8,0.2);
\coordinate (C) at (0,6);
\coordinate (D) at (6,12);
\coordinate (E) at (12,6);
\coordinate (F) at (6.6,0);
\coordinate (G) at (0.2,6);
\coordinate (H) at (6,11.6);
\coordinate (I) at (11.6,6);
\coordinate (J) at (6,0);
\coordinate (K) at (0,0);

\draw[dotted,thick,postaction={decorate}]
  (A) -- (B) -- (C) -- (D) -- (E) -- (F) -- (G) -- (H) -- (I) -- (J) -- (K);

\foreach \point in {(A), (1,2.9), (2.9,1), (3.1,6), (7,9), (9,7), (6,3)} {
  \fill \point circle[radius=3pt];

\node[blue] at (0,0.5) {(0,0)};
\node[blue] at (1.2,3.4) {(1,3)};
\node[blue] at (2.2,1.2) {(3,1)};
\node[blue] at (3.9,6) {(3,6)};
\node[blue] at (6.9,8.3) {(7,9)};
\node[blue] at (8.2,7) {(9,7)};  
\node[blue] at (6.3,2.4) {(6,3)};
}
\end{tikzpicture}
\end{scriptsize}
\end{minipage}
&
\begin{minipage}{0.5\textwidth}
\begin{scriptsize}
\begin{tikzpicture}[scale=0.5,
  decoration={
    markings,
    mark=between positions 0.1 and 1.0 step 0.2 with {\arrow[blue,line width=1mm]{stealth}}
  },
  rounded corners=5pt 
]

\coordinate (A) at (0,0);
\coordinate (B) at (3.1,0.9);
\coordinate (C) at (2.9,6.1);
\coordinate (D) at (7,9.2);
\coordinate (E) at (9.2,7);
\coordinate (F) at (6,2.8);
\coordinate (G) at (2.7,6.1);
\coordinate (H) at (7,9.3);
\coordinate (I) at (9.3,7);
\coordinate (J) at (6,2.7);
\coordinate (K) at (3.1,0.8);
\coordinate (L) at (0,-0.1);

\draw[postaction={decorate}]
  (A) -- (B) -- (C) -- (D) -- (E) -- (F) -- (G) -- (H) -- (I) -- (J) -- (K) -- (L);

\foreach \point in {(A), (1,2.9), (2.9,1), (3.1,6), (7,9), (9,7), (6,3)} {
  \fill \point circle[radius=3pt];

\node[blue] at (0,0.5) {(0,0)};
\node[blue] at (1.2,3.4) {(1,3)};
\node[blue] at (2.2,1.2) {(3,1)};
\node[blue] at (3.9,6) {(3,6)};
\node[blue] at (6.9,8.3) {(7,9)};
\node[blue] at (8.2,7) {(9,7)};  
\node[blue] at (6.3,2.4) {(6,3)};
}
\end{tikzpicture}
\end{scriptsize}
\end{minipage}
\end{tabular}
\caption{The dog's trail (left) and how the leash wraps around the lampposts (right).}
\label{dog}
\end{figure*}

\subsection{Quantitative results}
Table \ref{fig:results} summarizes the results of the various models. ``{\bf Correct}'' means that the model has given the correct answer and a correct solution, whereas ``{\bf Ans}'' means that the model has given the right answer but an erroneous solution. This could happen as some questions have the structure ``Is it possible to...'', where the model might answer ``No'' while providing the wrong motivation. All in all, 21 questions (5\%) were answered in this way, suggesting that it is essential not just to look at the final answer when evaluating the reasoning capabilities of models. There are also a few ``{\bf Sol}'' instances where the reasoning is correct and model has found the key idea, but makes a small calculation error leading to the wrong answer.

We see from table \ref{fig:results} that Mixtral8x7b is the worst-performing model, getting no solutions right, followed by Llama3.370B-versatile (10/50) and gpt-4o (14/50). The later models that have been trained with an explicit problem-solving objective \citet{gemini-2.0}, \citet{openaio1}, \citet{guo2025deepseek} fare much better, although there is still some variation.

\subsection{Spatial reasoning problems}

 This is a problem that confounded every model:
\begin{quote}
(Problem 11):  A dog is on an automatically retractable leash. If the owner is standing at (0,0) and the dog runs to (5,0), the extended part of the leach is 5 metres long, but when the dog returns to its owner at (0,0), the leach is rewinded and is 0 metres long again. However, if there is a lamppost at (1,3) and the dog runs from (0,0) to (5,0), then to (0,5) and then back to (0,0) again, the leash will loop around the lamppost so the extended part of the leash is now 2*sqrt(10), i.e. the distance from (0,0) to the lamppost and back again. Suppose now that there are lampposts at (1,3), (3,1), (6,3), (3,6), (9,7), and (7,9). The dog runs the following trail: (0,0) to (6,0) to (0,6) to (6,12) to (12,6) to (6,0) to (0,6) to (6,12) to (12,6) to (6,0) to (0,0). What is the length of the extended part of the leash when the dog has finished its run? Round the answer upwards to the closest integer.
\end{quote}

Figure \ref{dog} shows the dog's trail (left), and how the leash will wrap around the lampposts (right). This is an example of a problem which is easy to solve for a human (if allowed to use pen and paper to draw a figure), since the mathematics involved is just repeated use of the distance formula. Most adults would have intuitive idea of how a piece of string behaves when looped around some lampposts and then tightened, which makes it easy to come up with the picture in Figure \ref{dog}.

The reasoning errors committed by the models suggest that they cannot grasp the physics of the situation. {\bf o1} seemed to seize on the example in the question, and assumed that it should add the Euclidean distances from (0,0) to (some of) the lampposts and back again. {\bf deepseek-r1} comes to the same conclusion, even though in its reasoning printout (which is accessible for the user, unlike in the o1 and o3 models), deepseek seems to realize that the leash is wrapped twice around the diamond created by the four furthest lampposts, but fails to draw the right conclusion from that observation. {\bf o3-mini} explains (wrongly) that the dog is running one leap clockwise around the four furthest lampposts, and then counter-clockwise, meaning that ``the two windings cancel each other''. Somehow its conclusion is that the extented part of the leash is $2\sqrt{10}$, just as in the example in the question. The remaining models have non-sensical explanations.

Another problem where humans are helped by mental imagery is the following:
\begin{quote}
(Problem 19): Suppose you have two ordinary six-sided dice which you want to place on a wooden table so as few dots as possible are visible. The best way of doing this is placing them next to each other with the six dots facing downwards and the five dots facing each other. This way 2*(1+2+3+4)=20 dots will be visible altogether (the observer is allowed to walk around the table). We define v(n) to be the minimal number of dots visible on n dice placed on a table. You are given v(1)=15, v(2)=20, v(3)=26. What is v(37)?
\end{quote}

The correct answer is 95. The optimal placement is first to arrange 36 of the dice in a $6\times 6$ square, with ”1” facing upwards on each die, ”2” and ”3” facing outwards on the dice in the corners, and ”2” facing outwards on the dice along the edges, making 88 dots visible. The 37th die is placed with ”1” facing upwards, and its ”5” pressed against one of the ”3”s in the square of dice. The 37th die now exposes 1–4, but covers a ”3” which was previously visible. All in all, adding the 37th die will contribute an additional 7 visible dots, so v(37) = 88+7 = 95.

{\bf deepseek-r1} actually nailed this problem, giving essentially the explanation above, after an extensive chain-of-thought process (>22,000 tokens). {\bf o3-mini} realized the $6\times 6$ configuration, but then goes astray when placing the 37th die. {\bf o1} and {\bf gemini} instead suggested putting the dice in a line (which is sub-optimal), and also failed to correctly count the number of visible dots for that configuration. The remaining models tried to fit a numerical formula (e.g.\ a quadratic formula) based on the three examples given in the question, without considering the actual physics of the problem. These attempts all ended in failure.

Finally, we mention the following problem, which resulted in the largest number of incorrect solutions but correct answers:

\begin{quote}
    (Problem 26): We want to assign a number in $\{1 \ldots 12\}$ to each of the edges on a cube so that (1) each edge is assigned a different number, and (2) the sum of the four edges on one face of the cube will be the same for all faces. Determine whether this is possible or not. If it is possible, determine which number the edges on one face should add up to.
\end{quote}

A correct solution would first point out that each number would appear on two faces, meaning that the total number of numbers visible on the six faces is $2(1+\ldots +12) = 156$, which entails that the sum of each face is 156/6 = 26. All models except mixtral came this far. However, the second part of the solution is to show that there is a concrete assignment of numbers to edges that result in each face having the sum 26. {\bf deepseek-r1} tried to do this but came up with an erronous assignment. Only {\bf o3-mini} managed to get the solution completely correct. 

Several models concluded that assigning numbers to the edges as described in the question is possible just because twice the sum of 1..12 is divisible by 6, or equivalently that $1+\ldots +12$ is divisible by 3. But there are many sets of 12 numbers whose sum is divisible by 3 but which cannot be assigned to the edges of a cube in the way described in the question. The failure to realize this might have been due to the model having seen the problem in its training data (and knowing it to be solvable), or simply a failure to consider the physical constraints of the problem.

\subsection{Strategy problems}
Most LLMs were struggling with problems of a strategic nature. An example was the following:
\begin{quote}
(Problem 4): An ordinary tic-tac-toe board has 9 squares: (1,1) - (3,3). Now consider fric-frac-froe, which is played on an extended board where the top row has four squares (1,1)-(1,4), and the other two rows have three squares as before. The objective of fric-frac-froe is to have three markers in a row, just as in ordinary tic-tac-toe. Either find a winning strategy for the fric-frac-froe player who goes first, or explain why the game is a draw. 
\end{quote}

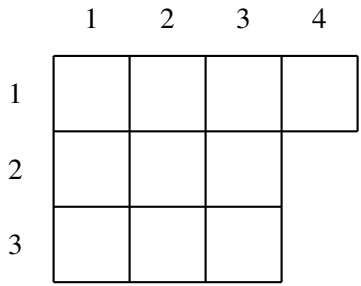
\begin{figure}[ht]
\begin{center}
\begin{tikzpicture}
    \draw[thick] (0,0) -- (0,3);
    \draw[thick] (1,0) -- (1,3);
    \draw[thick] (2,0) -- (2,3);
    \draw[thick] (3,0) -- (3,3);
    \draw[thick] (4,2) -- (4,3);
    \draw[thick] (0,0) -- (3,0); 
    \draw[thick] (0,1) -- (3,1); 
    \draw[thick] (0,2) -- (4,2);
    \draw[thick] (0,3) -- (4,3);
    \node at (-0.5, 0.5) {3};
    \node at (-0.5, 1.5) {2};
    \node at (-0.5, 2.5) {1};
    \node at (0.5, 3.5) {1};
    \node at (1.5, 3.5) {2};
    \node at (2.5, 3.5) {3};
    \node at (3.5, 3.5) {4};
\end{tikzpicture}
\caption{A fric-frac-froe board.}
\label{fric-frac-froe}
\end{center}
\end{figure}
That is, the board looks like Figure \ref{fric-frac-froe}. It does not take a human observer long to discover a winning strategy, which is to play (1,3). The second player has to block at (1,2), and now the correct move is (2,2). The second player has to block at (3,1), and now (3,3) wins, since both (1,1) and (2,3) are threatened and the second player cannot block both of them. (There are also winning strategies for the first player starting with (2,2)). 

{\bf o1}, {\bf o3-mini}, {\bf gemini}, and {\bf deepseek-r1} all correctly point out that the game is a win for the first player, but to back up that claim they either propose a strategy that would lead to a draw or even a loss for the first player, or gives an incomplete strategy. {\bf gemini-2.0-pro-exp} comes closest to a viable strategy by suggesting to start with (2,2) and follow up with (1,3). However, it proposes to play (1,3) also in the case where the second player plays (1,3) in the first move, which is clearly impossible. The remaining 4 models seem to find it obvious that the game is a draw, probably influenced by the many sources in their training material describing tic-tac-toe as being a draw.

A problem which is a combination of strategic and spatial reasoning is the following:
\begin{quote}
(Problem 30): A square-shaped swimming pool has its opposite corners at (0,0) and (2,2). A swimmer and a runner makes the following bet: They will start at the same time at (0,0). The goal of the swimmer is to swim to either (1,2) or (2,1). If he can reach either of those points and the runner is not already there when the swimmer arrives, the swimmer will win the bet, otherwise the runner will win the bet. Suppose the runner is twice as fast as the swimmer. Assuming both players are using their best strategy, who will win the bet? Explain the best strategies for the runner and the swimmer.
\end{quote}

\begin{figure}[ht]
\begin{center}
\begin{tikzpicture}[
  decoration={
    markings,
    mark=between positions 0.3 and 1.0 step 0.5 with {\arrow[blue,line width=1mm]{stealth}}
  },
  rounded corners=5pt 
]
    \coordinate (A) at (0.1,0.1);
    \coordinate (B) at (2,2);
    \coordinate (C) at (4,2);
    \draw[thick,fill=blue!10] (0,0) rectangle (4,4);
    \draw[postaction={decorate}, dashed] (A) -- (B) -- (C);
    \node at (2.2,1.7) {S1};
    \node at (1.6,4.1) {$\bullet$};
    \node at (1.6,4.4) {R1};
    \node at (4.4,2) {S2};
    \node at (4.5,2.4) {R2};
    \node at (4.1,2.4) {$\bullet$};
    \node at (4,2) {$\bullet$};
    \node at (2,2) {$\bullet$};
    \coordinate (D) at (-0.1, 0);
    \coordinate (E) at (-0.1, 4.1);
    \coordinate (F) at (1.6, 4.1);
    \draw[postaction={decorate}, dotted] (D) -- (E) -- (F);
    \coordinate (G) at (1.6,4.1);
    \coordinate (H) at (4.1,4.1);
    \coordinate (I) at (4.1, 2.4);
    \draw[postaction={decorate}, dotted] (G) -- (H) -- (I);
    \node at (-0.4,0) {0};
    \node at (-0.4,2) {1};
    \node at (-0.4,4) {2};
    \node at (0,-0.3) {0};
    \node at (2,-0.3) {1};
    \node at (4,-0.3) {2};
\end{tikzpicture}
\caption{The swimming pool of problem 30, showing the swimmer's route (dashed line) and the runner's route (dotted line).}
\label{pool}
\end{center}
\end{figure}
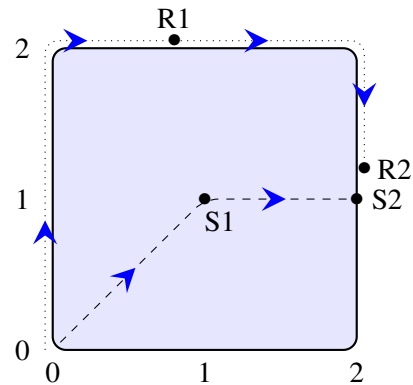
The answer is that swimmer will win the bet. He has to use the right strategy, though. If the swimmer aims for (1,2) and swims there in a straight line, he will cover the distance in $\sqrt{5}=2.23$ time units. However, the runner will already be there on his arrival, since the distance along the perimeter is~3, which will only take the runner 1.5 time units. A better swimming strategy is to aim for the centre of the pool at (1,1), as illustrated in Figure~\ref{pool}. The runner, who cannot be sure which point the swimmer is aiming for, has to commit to start running either along the x-axis or the y-axis (the runner has no better strategy than randomly selecting between these). Let's say he starts running along the y-axis. When the swimmer reaches position S1 at (1,1) after $\sqrt{2}$=1.4 time units, the runner has reached position R1 at (0.8, 2). The swimmer now turns slightly right and aims for position S2 at (2,1), which he will reach in 1 additional time unit. However, the runner can only reach position R2 at (2,1.2) in that time, so the swimmer will win the bet.

The only model to completely solve the problem was {\bf o1}, who realized the key idea that the swimmer can change direction in the middle of the pool. {\bf deepseek-r1} concludes its output by the self-contradictory statement ``{\em the swimmer will win the bet by using their best strategy to randomize between the targets, ensuring a 50\% chance of winning}''. The other models all suggest that the runner will win, focusing their explanations on the fact that the runner is quicker.

No model could solve problem 22:
\begin{quote}
     (Problem 22): I have a collection of 5 triangles, T1 to T5. All the interior angles (when expressed as degrees) of T1 to T5 are distinct and are found in the set \{1,2,3,4,5,6,7,8,9,10,167,168,169, 170,171\}. You can now point to a number in A and ask an oracle which triangle this angle belongs to. What is the minimal number of such questions you have to ask before you are guaranteed to know which triangle each of the 15 numbers in A belongs to? 
\end{quote}
The answer, which requires careful a careful analysis, is that 6 questions to the oracle are sufficient and also necessary in the worst case to know which angle is in which triangle  (see Appendix \ref{appendix:problem22} for a complete solution). 

Since the angles in a triangle must add up to 180 degrees, the 5 bigger angles 167--171 must be in different triangles. All the models have realized this, but they propose various incorrect numbers as the solution. {\bf o3-mini} suggests (incorrectly) that it is necessary to ask the oracle about the triangle of each of 5 largest angles, plus an additional 5 questions for 5 or smaller angles, all in all 10 questions (the location of the 5 remaining angles can be deduced from the fact that the angles in a triangle must add up to 180 degrees). {\bf deepseek-r1} improves on this suggestion by pointing out (correctly) that it is sufficient to ask about the location of 4 of the large angles and 4 of the smaller angles, since the angles of the fifth triangle can be deduced by the process of elimination. The strategies proposed by o3-mini and deepseek-r1 would both lead to the desired information, but they are not optimal in terms of the number of questions. 

{\bf o1} starts off very well by observing (correctly) that there are 8 possible groupings of the angles in A into triples that add up to 180 degrees, and because you need to assign a name T1...T5 to each of the triangles, there are all in all $5!\cdot 8=960$ possibilities. o1 also concludes, by an information-theoretic argument, that at least 5 questions to the oracle are necessary. Unfortunately, it does not consider whether 5 questions also are sufficient, but simply states (wrongly) that the necessary information can be obtained with 5 questions using ``a suitably clever choice of which angles to query'', and does not specify what that clever choice would be. o1's proposed solution is therefore incorrect.

{\bf llama3.3}, {\bf mixtral}, {\bf gemini} and {\bf o1-preview} suggest respectively that 3, 4, 5, and 7 questions to the oracle are necessary and sufficient, either accompanied by an incorrect question strategy or a general argument without any concrete strategy.

\subsection{Numerical and set-theoretic problems}
Perhaps not surprisingly, the models are stronger at purely mathematical problems without any strategical element or any need for spatial reasoning. However, problem 17 proved too difficult for all models:
\begin{quote}
    (Problem 17): We write the prime numbers in six columns, as follows: In the first row, we write the first six primes: 2,3,5,7,11,13. In the second row, we write the next six primes in reverse order: 37,31,29,23,19,17. We keep on alternating between writing the next six primes in order on odd-numbered rows, and the next six primes in reverse order on even-numbered rows. After a new row has been added, we compute the sums of all the columns. After how many rows will we see (for the first time) that the third column has the largest sum of all columns? Either answer with a number, or explain why this will never happen.
\end{quote}

The correct answer is that this will first happen on line 83. However, all models claimed that it will never happen that the third column will have the largest value, citing numerical evidence from studying the first couple of lines of the constructed prime table. It is notoriously hard to reason about the additive properties of prime numbers, so perhaps the only viable to solve this problem in a reasonable amount of time is to write a small program and execute it, which no model attempted. 

A far simpler problem that proved to be surprisingly difficult was this one:
\begin{quote}
    (Problem 7): We have two disjoint sets of numbers: $A$, with $n$ members, and $B$ with $n+1$ members. We want to construct a sequence of numbers which is $2n+1$ numbers long, and every second number is selected from $A$ and every second number from $B$, and the sequence has to start and end with a number from $A$. Either suggest a method for doing this, or explain why such a method cannot exist.
\end{quote}
A possible general method is ``always select the smallest member from $A$ and the smallest member from $B$''. As a simple example (not given to the models), consider $A=\{1\}$ and $B=\{2,3\}$, where the proposed method would produce the  sequence $1,2,1$. It is trivial to see that the proposed method would always work; however, all the models except {\bf o1} and {\bf o3-mini} either claimed that the problem does not have a solution or gave solutions that were completely wrong, because the models all presuppose that each element could only be selected once from each set. We surmise that problem 7 is similar to some problem used in the models' training material where the select-only-once criterion is essential. If we put the problem in a slightly different context\footnote{The alternative formulation was: ``{\em The proportion of winning lottery tickets in three different lotteries A, B, and C can be found in this set: $\{0.1, 0.05\}$. Give an example of what the winning chances might be in the three lotteries.}''}, most models could solve it correctly.

Another situation where strong prior assumptions seem to lead the models astray was the following:
\begin{quote}
    (Problem 47): We will call a set of positive integers ``progressive'' if at least three of the numbers in the set belong to an arithmetic progression with a common difference larger than 1. For example, a set containing 2,6,42 is progressive, since these three numbers belong to the same arithmetic progression 2,4,6,8,...,40,42,... Either construct a set of of positive integers which is not progressive and has at least 5 members, or explain why no such set exists.
\end{quote}
The correct answer is that no such set exists, as every non-progressive set can contain at most 2 odd and 2 even numbers. However, only {\bf o1} and {\bf o3-mini} realized this. All the remaining models claimed the contrary, and stated examples like $\{1,2,4,8,16\}$, which was claimed to be non-progressive with the motivation ``{\em the differences between consecutive elements are not equal}''. We found this somewhat surprising, since the prompt even contained a similar example with an explanation of why the set indeed is progressive. We can only assume that the pre-training probability distribution of the model strongly leads it to its answer, ignoring the example in the prompt.

\section{Discussion}
Throughout the erroneous answers to the 50 example problems, we can see many traits we also see in many human math and engineering students failing to solve similar problems:
\begin{itemize}
\item Making arithmetic errors 
\item Disregarding constraints in the question formulation (as several models did for problem 47 above)
\item Adding unwarranted assumptions (as in problem 7 above)
\item Over-reliance on preliminary numerical evidence, as in problem 17
\item Trying to shoe-horn a problem into a known solution method, as in the ``fric-frac-froe'' game, where several models seemed to assume the game to be a draw, just like tic-tac-toe.  
\item Failure to find a key idea (the problem is just too difficult).
\end{itemize}
However, the failure to add unstated but common-sense assumptions is rather unique to models, e.g.\ that it is impossible for the runner to run in a swimming pool (this running strategy was suggested by {\bf o1-preview} in problem 30). 

On the other hand, in particular the latest models {\bf o1, o3-mini, deepseek-r1} and {\bf gemini} seem to have a large base of mathematical knowledge. Throughout the solutions, we could see the models make reference to Pick's theorem, the Frobenius coin problem, and Eulerian circuits, among others. Even though we have focused on faulty reasoning in this paper, state-of-the-art models (in particular the o1, o3, and deepseek models) have impressive reasoning capabilities and can solve quite difficult problems. The problem, as always with LLMs, is that also the erroneous solutions can look good at a cursory inspection, in particular if the reader has limited mathematical knowledge.  

\section{Limitations}
The results presented in this article provide a snapshot of the mathematical abilities of some state-of-the-art large language models (LLMs) in early 2025. The article provide some insights to the blind spots and shortcomings of LLMs when it comes to mathematical reasoning, but is unclear how much one can generalize from the results, due to the following:
\begin{itemize}
    \item LLM technology is developing rapidly, and it is perfectly possible that state-of-the-art models can solve more problems than described here just a few months from now.
    \item Each model was just queried once, due to time constraints (each solution was assessed manually, which took considerable amounts of time). It is possible that in some cases, a model might have produced a better answer in a second or third try.
    \item We do not have access to the internals of the systems, in particular, we could not scrutinize the chain-of-thought printouts from the {\bf o1}, {\bf o1-preview}, {\bf o3-mini}, and {\bf gemini-2.0-pro-exp} models. \item 8 state-of-the-art models were tested, but there are of course more models than these, and the models also exist in several versions. Due to time contraints, we could not try all of them.
    \item The 50 problems in the problem set only covered certain sub-areas of high school mathematics. Notably, trigonometry and calculus were missing.
    \item Though we strived to invent original problems which would not appear in the training set of any model, our imagination is limited, and it is perfectly possible that some model had seen some problem (or something very similar) in its training phase.
\end{itemize}

\bibliography{custom}

\newpage 

\appendix

\section{Appendix: All 50 problems}

\begin{enumerate}

    \item Let $n_i$ be the numeral obtained by writing the number 97 in base $i$. Then interpret $n_2, \ldots, n_9$ as decimal numbers, and let $s$ be the sum of those numbers. What is $s$ modulo 97 (in base 10)?
    
    \vspace{.2cm}
    
    
    \item  We have a calculator that respects the ordinary laws of arithmetic precedence (e.g., 2+3*4 will result in 14). We now randomly press (with a uniform probability) one of the digits 0--9, then either '+' or '*', then another random digit, then then either '+' or '*' again, and then another random digit. Finally, we press '=', and note down the answer. If we keep repeating this experiment over and over, what is the expected average result?
    
    \vspace{.2cm}
    
    
    \item  On a black-and-white computer screen, digits and numbers are displayed as bitmaps with 7 rows and 5 columns. For instance, an "I" is displayed like this:
    \begin{verbatim}
        01110
        00100
        00100
        00100
        00100
        00100
        01110
    \end{verbatim} 
    The bitmap for "I" contains 3 ones on the first row, 1 one on the second row, etc., that is, [3,1,1,1,1,1,3] ones, counting from the first row to the last.  Which letter in A--Z has this number of ones: [4,2,2,4,2,2,4], counting from first row to the last?
    
    \vspace{.2cm}
    

    \item An ordinary tic-tac-toe board has 9 squares: (1,1) - (3,3). Now consider fric-frac-froe, which is played on an extended board where the top row has four squares (1,1)-(1,4), and the other two rows have three squares as before. The objective of fric-frac-froe is to have three markers in a row, just as in ordinary tic-tac-toe. Either find a winning strategy for the fric-frac-froe player who goes first, or explain why the game is a draw. 
    
    \vspace{.2cm}
    

    \item We will call a binary tree with numbers at each node a 'labeled binary tree'. Either give an example of a labeled binary tree of depth 3 whose pre-order traversal and post-order traversal yields the same sequence of numbers, or explain why no such tree can exist.

    \vspace{.2cm}


    \item A rectangle has sides with non-zero integer lengths. Adding the length of the perimeter and the area of the rectangle yields 9793. How long are the sides?

    \vspace{.2cm}


    \item We have two disjoint sets of numbers: A, with n members, and B with n+1 members. We want to construct a sequence of numbers which is 2n+1 numbers long, and every second number is selected from A and every second number from B, and the sequence has to start and end with a number from A. Either suggest a method for doing this, or explain why such a method cannot exist.

    \vspace{.2cm}

    
    \item We have 4 points in the plane: p1, p2, p3, p4, and construct a polygon by drawing a line from p1 to p2, from p2 to p3, from p3 to p4, and from p4 back to p0 again. Suppose p1=(3,4), p2=(7,7), and p3=(10,3). If we want the polygon to be a square, where should p4 lie? Either give the coordinates of p4, or explain why no such point can exist.
    
    \vspace{.2cm}

    
    \item Let $a_0$ be the factorial of $1000^{1000}$, and let ${a_k}$ be the sum of digits in $a_{k-1}$, for $k>0$. After $i$ steps, $a_i, a_{i+1}, a_{i+2}, \ldots$ will be same number, which is a single digit. Which digit?

    \vspace{.2cm}


    \item Let $x$ be a positive integer and define the following rule $f$: $f(x) = x/3$ if $x$ is divisible by 3, otherwise $f(x) = 2x+1$. We are interested in how many times we must apply this rule before we reach the number 1. For $x=4$, we need 3 applications: $f(4)=9, f(9)=3, f(3)=1$. Let us use the notation $g(x)$ to denote the smallest $i$ such that $i$ applications of $f$ starting from $x$ results in 1. As we saw, $g(4)=3$. If no such $i$ exists, we let $g(x) = -1$. What is $g(1) + g(2) + \ldots + g(100)$?

    \vspace{.2cm}


    \item  A dog is on an automatically retractable leash. If the owner is standing at (0,0) and the dog runs to (5,0), the extended part of the leach is 5 metres long, but when the dog returns to its owner at (0,0), the leach is rewinded and is 0 metres long again. However, if there is a lamppost at (1,3) and the dog runs from (0,0) to (5,0), then to (0,5) and then back to (0,0) again, the leash will loop around the lamppost so the extended part of the leash is now 2*sqrt(10), i.e. the distance from (0,0) to the lamppost and back again. Suppose now that there are lampposts at (1,3), (3,1), (6,3), (3,6), (9,7), and (7,9). The dog runs the following trail: (0,0) to (6,0) to (0,6) to (6,12) to (12,6) to (6,0) to (0,6) to (6,12) to (12,6) to (6,0) to (0,0). What is the length of the extended part of the leash when the dog has finished its run? Round the answer upwards to the closest integer.

    \vspace{.2cm}  


    \item We have a convex polygon in the plane, with vertices in $(3,0)$, $(1, 2.5)$, $(8, 9.8)$, $(12, 8.5)$, and $(11, -0.5)$. How many points with integer coordinates are contained in this polygon (not counting those on the perimeter)?

    \vspace{.2cm}

    
    \item Suppose you randomly remove 15 paper sheets from a book. Each sheet has a page number written on either side of the sheet. Can these page numbers add up to 2000? Explain how you reached your conclusion.

    \vspace{.2cm}
    

    \item We have a rectangular pool table with near left corner in (0,0), and the far right corner in (5,11). A ball is sent off from the near left corner in the direction (1,1). How many times will the ball bounce off a wall before ending up in a corner? Assume that the incoming angle is equal to the outgoing angle at each bounce.

    \vspace{.2cm}
    

    \item I have fifteen dice that I want to place on a flat empty wooden surface in such a way that as many dice as possible will have all six faces concealed to an observer. The observer is allowed to walk around the table but not to touch the dice. Determine the maximum number of dice you can conceal, and explain how to best place the dice.  

    \vspace{.2cm}


    \item Suppose we have a ordinary clock with an hour hand and a minute hand. We are interested in the angle between the hands measured from the minute hand clockwise to the hour hand. For example, the angle is 30 degrees at 1pm. At how many occasions from 1pm to 2pm (inclusive) will the angle between the hands be an integer?
       
    \vspace{.2cm}


    \item We write the prime numbers in six columns, as follows: In the first row, we write the first six primes: 2,3,5,7,11,13. In the second row, we write the next six primes in reverse order: 37,31,29,23,19,17. We keep on alternating between writing the next six primes in order on odd-numbered rows, and the next six primes in reverse order on even-numbered rows. After a new row has been added, we compute the sum of each column. After how many rows will we see (for the first time) that the third column has the largest sum of all columns? Either answer with a number, or explain why this will never happen.

    \vspace{.2cm}
    

    \item We write the powers of 2 in five columns, as follows: In the first row, we write the first five powers of 2: 1,2,4,8,16. In the second row, we write the next five powers of 2 in reverse order: 512,256,128,64,32. We keep on alternating between writing the next five powers of two in order on odd-numbered rows, and the next five powers of two in reverse order on even-numbered rows. After a new row has been added, we compute the sum of each column. After how many rows will we see (for the first time) that the second column has the largest sum of all columns? Either answer with a number, or explain why this will never happen.

    \vspace{.2cm}
    

    \item Suppose you have two ordinary six-sided dice which you want to place on a wooden table so as few dots as possible are visible. The best way of doing this is placing them next to each other with the six dots facing downwards and the five dots facing each other. This way 2*(1+2+3+4)=20 dots will be visible altogether (the observer is allowed to walk around the table). We define v(n) to be the minimal number of dots visible on n dice placed on a table. You are given v(1)=15, v(2)=20, v(3)=26. What is v(37)?

    \vspace{.2cm}
    

     \item Let us call a positive integer "good" if it has a 7 somewhere in its decimal representation. How many of the first 10 million positive integers are good?

     \vspace{.2cm}


    \item I have six paper slips which have numbers written on them: respectively 2,2,3,3,5, and 7. I want to select a subset of the paper slips and compute the sum of those numbers. Which is the smallest integer larger than 1 which cannot be obtained this way?

    \vspace{.2cm}


    \item I have a collection of 5 triangles, T1 to T5. All the interior angles (when expressed as degrees) of T1 to T5 are distinct and are found in the set A = \{1,2,3,4,5,6,7,8,9,10,167,168,169 ,170,171\}. You can now point to a number in A and ask an oracle which triangle this angle belongs to. What is the minimal number of such questions you have to ask before you are guaranteed to know which triangle each of the 15 numbers in A belongs to?

    \vspace{.2cm}
    

    \item I have the points (0,1), (3,2), and (4,1), and construct a circle where the three points lie on the perimeter. If we call the centre of the circle (xc, yc), what can you say about xc relative to the x-coordinates of the three points: is it smaller, equal, or larger? What about yc relative to the y-coordinates of the three points?

    \vspace{.2cm}


    \item Suppose we sort all numbers from 1 to 1000 (inclusive) in an ascending order according the digit sum of the number. 501 would appear before 129 in such a sorted list, since 5+0+1 is less than 1+2+9. If i and j have the same digit sum, but i is smaller than j, then i should appear before j in the sorted list. At which position in the list would you find the number 721? 

    \vspace{.2cm}   
    
    
    \item I have a rectangular grid of 80x80 identical squares, and a number of tiles that can be placed on the grid. The tiles come in two varieties, both varieties covering 4 squares: 2x2-tiles and 1x4-tiles. The 1x4 tiles can be placed either horisontally or vertically. Is it possible to tile the grid with 799 2x2 tiles and 801 1x4-tiles? If it is possible, suggest a method. If it is not possible, explain why not.
    
    \vspace{.2cm}
    

    \item We want to assign a number in $\{1 \ldots 12\}$ to each of the edges on a cube so that (1) each edge is assigned a different number, and (2) the sum of the four edges on one face of the cube will be the same for all faces. Determine whether this is possible or not. If it is possible, determine which number the edges on one face should add up to.

    \vspace{.2cm}
    

    \item Divide the area from (0,0) to (3,3) into 9 equally large squares. We now want to fill in the perimeter of these 9 squares without lifting the pen. What is the minimal length of the line you need to draw?

    \vspace{.2cm}


    \item We place the numbers 1--9 in a 3x3 grid, and add all the numbers on the rows, the columns, and the two diagonals to form one big sum. The obtained sum will depend on how we placed the numbers in the grid. What is the difference between the largest and the smallest sums obtainable in this way?
    
    \vspace{.2cm}
    
    
    \item Some integers can be written as 6x+7y, where x and y are non-negative integers, but there is a largest integer N that cannot be written this way. Find N, and explain why all numbers larger than N can be written as 6x+7y.

    \vspace{.2cm}

    
    \item A square-shaped swimming pool has its opposite corners at (0,0) and (2,2). A swimmer and a runner makes the following bet: They will start at the same time at (0,0). The goal of the swimmer is to swim to either (1,2) or (2,1). If he can reach either of those points and the runner is not already there when the swimmer arrives, the swimmer will win the bet, otherwise the runner will win the bet. Suppose the runner is twice as fast as the swimmer. Assuming both players are using their best strategy, who will win the bet? Explain the best strategies for the runner and the swimmer.
    
    \vspace{.2cm}
    

    \item We put the numbers 1--32 in a random order and call this sequence S. We now want generate 33 permutations $T_0$--$T_{32}$ of the numbers 1--32 so that $T_{i}$ has the same number as S at exactly $i$ positions. For instance, S and $T_1$ might both have the number 17 in position 22, but not overlap in any other positions. Either describe a method of how to generate $T_0$--$T_{32}$, or explain why it is impossible.

    \vspace{.2cm}
    

    \item You have a collection of 8 line segments, which have the length of the first 8 odd numbers. Can you construct a square using each of these line segments exactly once?

    \vspace{.2cm}


    \item You have a collection of 10 line segments, which have the length of the first 10 odd numbers. Can you construct a square using each of these line segments exactly once?

    \vspace{.2cm}


    \item You have a square with the dimensions 100x100 metres, and want to place a rectangle 10 metres wide and y metres long on top of the square so that no piece of the rectangle extends beyond the borders of the square. What is the maximum length y possible for the rectangle? Round the answer downwards to an integer.   
    
    \vspace{.2cm}


     \item A pandigital number contains all digits 0--9 at least once. What percentage of all 10-digit pandigital numbers are divisible by~3?

    \vspace{.2cm}


     \item A pandigital number contains all digits 0--9 at least once. What percentage of all 11-digit pandigital numbers are divisible by~3?

    \vspace{.2cm}


    \item We have an empty screen whose lower left corner is at (0,0) and the upper right corner at (1000,1000). We open 5 windows on the screen: 

    \begin{tabular}{ccc}
    Window number & Lower left & Upper right\\
    1 & (100,200) & (900,400) \\
    2 & (200,100) & (700,900) \\
    3 & (0, 500) & (400,800) \\
    4 &  (300,200) & (800,800) \\
    5 & (600,100) & (1000,500)
    \end{tabular}

    How large a proportion of the screen background will still be visible after having opened the windows above?
    
    \vspace{.2cm}
        
    
    \item Find a positive integer $n$ such that the sum of the digits in $n^2$ is 101, or explain why no such $n$ can exist.
    
    \vspace{.2cm}
    

    \item In how many ways can you tile a 10x2 grid with 1x2 dominoes? (We assume that all dominoes are blank and indistinguihable).

    \vspace{.2cm}
    
     
    \item We have a list of 100 numbers sort in ascending order, but we want the list sorted in descending order. The only operation at our disposal is to swap to numbers at position $k$ and $k+2$ in the list. Determine the number of such swap operations necessary to get the list sorted in descending order, or explain why it is not possible at all.

    \vspace{.2cm}   
    
    
    \item A man is climbing a staircase with 10 steps, numbered 1--10. Each second, the man climbs one step with probability 0.5, or descends one step with probability 0.5. If the man is at the bottom of the stairs (step 0), then obviously he cannot descend further, so in that case he stays put with probability 0.5, or climbs one step with probability 0.5. If the man starts at step 0, which is the expected step he will be at after 10 seconds? Round to the nearest integer. 

    \vspace{.2cm}


    \item We select a sequence of 10 random digits with repetition (i.e., the same digit can be chosen more than once) using a uniform distribution. What is the probability that the product of the digits in the sequence is odd?
    
    \vspace{.2cm}   
    

    \item We select a sequence of 10 random digits with repetition (i.e., the same digit can be chosen more than once) using a uniform distribution. What is the probability that the sum of the digits in the sequence is odd?
    
    \vspace{.2cm}   
    

    \item We have a calculator that respects the ordinary laws of arithmetic precedence (e.g., 2+3*4 will result in 14). We now randomly press (with a uniform probability) one of the digits 0--9, then either '+' or '*', then another random digit, then then either '+' or '*' again, and then another random digit. Finally, we press '='. What is the probability that the result is odd?   

    \vspace{.2cm}  
    
 
    \item Consider the set of strings matching the regular expression "a+b+a+". How many strings of length 100 match this regular expression? Only count the cases where the whole string matches the regular expression.  

    \vspace{.2cm}
    

    \item You have a cardboard cylinder whose outer diameter is 100~mm. You roll a paper which is 100,000~mm long and 1~mm thick on the cylinder. How many times do you need to rotate the cylinder a full 360 degrees before you have rolled up the whole paper? Answer with the nearest higher integer.

    \vspace{.2cm}
    

    \item We will call a set of positive integers ``progressive'' if at least three of the numbers in the set belong to an arithmetic progression with a common difference larger than 1. For example, a set containing 2,6,42 is progressive, since these three numbers belong to the same arithmetic progression 2,4,6,8,...,40,42,... Either construct a set of of positive integers which is not progressive and has least 5 members, or explain why no such set exists.
    
    \vspace{.2cm}
    

    \item Can the numbers 1--25 be placed in different groups such that the product of the numbers in each group is the same? Explain.

    \vspace{.2cm}

    
    \item Suppose $n = a_1 + a_2 + \ldots a_k$, and consider the claim "n is divisible by d if and only if each $a_i$ is divisible by d". Is (a) the "if" part the claim guaranteed to be true but not the "only if" part, or (b) is it the other way around, or (c) are both guaranteed to be true, or (d) is neither? Explain.
    
    \vspace{.2cm}
    

    \item Anna and Bert is playing the following game: First, a positive integer less than 1000 is randomly generated. Anna goes first and chooses to subtract either 1 or 2 from the number. Then Bert can choose to subtract either 1 or 2 from the resulting number. Then it's Anna's turn again, and the players keep alternating, subtracting either 1 or 2. The player who reaches negative infinity wins. Is there a winning strategy for either Anna or Bert? Explain. 
    
\end{enumerate}

\section{Appendix: Solution to problem 22}
\label{appendix:problem22}

The answer is that 6 questions are sufficient and also necessary in the worst case. Careful analysis of the problem reveals that there are 8 different possible configurations, called {\bf A}-{\bf H} (see Figure \ref{fig:problem22})):

    \vspace{.2cm}

    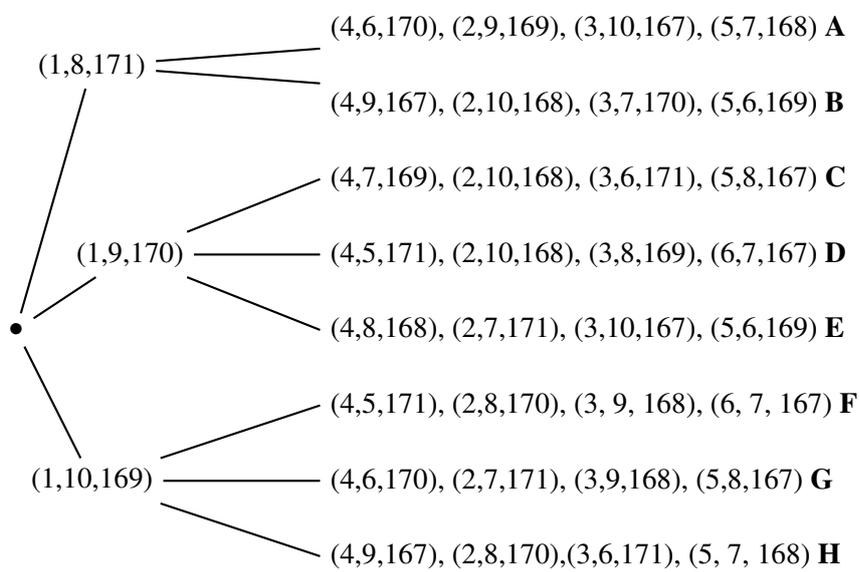
\begin{figure*}[b]
        \begin{tikzpicture}
            \node at (0,3) (root) {$\bullet$};
            \node at (1,6.5) (18) {(1,8,171)};
            \node at (1.5,4) (19) {(1,9,170)};
            \node at (1,1) (110) {(1,10,169)};
            \node[anchor=west] at (4,7) (181) {(4,6,170), (2,9,169), (3,10,167), (5,7,168) {\bf A}};
            \node[anchor=west] at (4,6) (182) {(4,9,167), (2,10,168), (3,7,170), (5,6,169) {\bf B}};
            \node[anchor=west] at (4,5) (191) {(4,7,169), (2,10,168), (3,6,171), (5,8,167) {\bf C}};
            \node[anchor=west] at (4,4) (192) {(4,5,171), (2,10,168), (3,8,169), (6,7,167) {\bf D}};
            \node[anchor=west] at (4,3) (193) {(4,8,168), (2,7,171), (3,10,167), (5,6,169) {\bf E}};
            \node[anchor=west] at (4,2) (101) {(4,5,171), (2,8,170), (3, 9, 168), (6, 7, 167) {\bf F}};
            \node[anchor=west] at (4,1) (102) {(4,6,170), (2,7,171), (3,9,168), (5,8,167) {\bf G}};
            \node[anchor=west] at (4,0) (103) {(4,9,167), (2,8,170),(3,6,171), (5, 7, 168) {\bf H}};
            \draw[thick] (root) -- (18);
            \draw[thick] (root) -- (19);
            \draw[thick] (root) -- (110);
            \draw[thick] (18) -- (182);
            \draw[thick] (18) -- (181);
            \draw[thick] (19) -- (4,5);
            \draw[thick] (19) -- (192);
            \draw[thick] (19) -- (4,3);
            \draw[thick] (110) -- (4,2);
            \draw[thick] (110) -- (102);
            \draw[thick] (110) -- (4,0);
        \end{tikzpicture}
    \caption{The possible triangles of problem 22.}
    \label{fig:problem22}
    \end{figure*}
    
    \vspace{.2cm}

    In addition, the triangles might have different ID numbers, e.g.\ (1,8,171) might be any of T1 -- T5, so the total number of possibilities are $5!\cdot 8 = 960$.

    The stategy for querying the oracle is described visually in Figure \ref{fig:problem22}. First ask the oracle about about 1 and 8. If 1 and 8 are in the same triangle, ask about 4 and 6. If they are in the same triangle, solution {\bf A} is correct, and you know the ID numbers of two of the triangles. To get the remaining ID numbers, ask about 2 and 3, and by process of elimination one can also conclude the ID of the triangle containing 5. (This took 6 questions).

    If 4 and 6 are not in the same triangle, solution {\bf B} is correct, and you know the ID numbers of the triangles containing 1, 4 and 6. Then ask about 2, and you will have all necessary information. (This took 5 questions).

    If the first two questions reveal that 1 and 8 are not in the same triangle, ask about 9. If 1 and 9 are in the same triangle, one of solutions {\bf C, D, E} is correct, otherwise one of {\bf F, G, H} is correct. In either case, asking about 2,6, and 3 will give all the necessary information (totally 6 questions).

\section{Appendix: Detailed results}

Table \ref{fig:results} details the results of the various models. Each problem has two entries for each model: A green checkmark in the leftmost position indicates that the model got the right answer to the question, and a checkmark in the rightmost position means that the solution is correctly motivated. For the most part, models get either both or none of these right, but there are a number of instances when the model gives a correct answer but an erroneous motivation. This suggests that it is essential not just to look at the final answer when evaluating the reasoning capabilities of models.

\begin{figure*}[p]
\begin{center}
\begin{tiny}
\begin{tabular}{r|cc|cc|cc|cc|cc|cc|cc|cc|cc|}
  \# &
  \multicolumn{2}{|c|}{\parbox{1.2cm}{Mixtral\\8x7b}} & 
  \multicolumn{2}{c|}{\parbox{1.2cm}{Llama\\3.3-70B\\versatile}} & 
  \multicolumn{2}{c|}{\parbox{1.2cm}{Gemini\\2.0-pro-exp}} &
  \multicolumn{2}{c|}{\parbox{1.2cm}{gpt-4o}} &
  \multicolumn{2}{c|}{\parbox{1.2cm}{o1-preview}}&
  \multicolumn{2}{c|}{\parbox{1.2cm}{o1}}&
    \multicolumn{2}{c|}{\parbox{1.2cm}{o3-mini}}&
    \multicolumn{2}{c|}{\parbox{1.2cm}{deepseek-r1}}\\
  \hline

1 & \no & \no & \no & \no & \ok & \ok & \no & \no & \ok & \ok & \ok & \ok & \ok & \ok & \ok & \ok\\
2 & \no & \no & \no & \no & \ok & \ok & \no & \hm & \ok & \ok & \ok & \ok & \ok & \ok & \ok & \ok\\
3 & \no & \no & \no & \no & \no & \no & \no & \no & \ok & \ok & \ok & \ok & \ok & \ok & \ok & \ok\\
4 & \no & \no & \no & \no & \hm & \no & \no & \no & \no & \no & \hm & \no & \hm & \no & \hm & \no\\
5 & \no & \no & \no & \no & \no & \no & \no & \no & \ok & \ok & \no & \no & \ok & \ok & \ok & \ok\\
6 & \no & \no & \ok & \ok & \ok & \ok & \no & \no & \ok & \ok & \ok & \ok & \ok & \ok & \ok & \ok\\
7 & \no & \no & \no & \no & \no & \no & \no & \no & \no & \no & \no & \no & \ok & \ok & \ok & \ok\\
8 & \no & \no & \no & \no & \ok & \ok & \ok & \ok & \no & \no & \ok & \ok & \ok & \ok & \ok & \ok\\
9 & \no & \no & \no & \no & \ok & \ok & \ok & \ok & \ok & \ok & \ok & \ok & \ok & \ok & \ok & \ok\\
10 & \no & \no & \no & \no & \no & \no & \no & \no & \no & \no & \no & \hm & \no & \no & \no & \no\\
11 & \no & \no & \no & \no & \no & \no & \no & \no & \no & \no & \no & \no & \no & \no & \no & \no\\
12 & \no & \no & \no & \no & \no & \no & \no & \no & \no & \no & \ok & \ok & \ok & \ok & \ok & \ok\\
13 & \hm & \no & \no & \no & \ok & \ok & \no & \no & \ok & \ok & \ok & \ok & \ok & \ok & \hm & \no\\
14 & \no & \no & \no & \no & \ok & \ok & \ok & \ok & \ok & \ok & \ok & \ok & \ok & \ok & \ok & \ok\\
15 & \no & \no & \no & \no & \no & \no & \no & \no & \no & \no & \no & \no & \ok & \ok & \ok & \ok\\
16 & \no & \no & \no & \no & \no & \no & \no & \no & \no & \no & \ok & \ok & \ok & \ok & \ok & \ok\\
17 & \no & \no & \no & \no & \no & \no & \no & \no & \no & \no & \no & \no & \no & \no & \no & \no\\
18 & \no & \no & \hm & \no & \no & \no & \hm & \no & \hm & \no & \hm & \no & \ok & \ok & \ok & \ok\\
19 & \no & \no & \no & \no & \no & \no & \no & \no & \no & \no & \no & \no & \no & \no & \ok & \ok\\
20 & \no & \no & \no & \no & \ok & \ok & \no & \hm & \ok & \ok & \ok & \ok & \ok & \ok & \ok & \ok\\
21 & \no & \no & \no & \no & \ok & \ok & \ok & \ok & \ok & \ok & \ok & \ok & \ok & \ok & \ok & \ok\\
22 & \no & \no & \no & \no & \no & \no & \no & \no & \no & \no & \no & \no & \no & \no & \no & \no\\
23 & \no & \no & \no & \no & \ok & \ok & \no & \no & \ok & \ok & \ok & \ok & \ok & \ok & \ok & \ok\\
24 & \no & \no & \no & \no & \no & \hm & \no & \no & \no & \hm & \ok & \ok & \ok & \ok & \ok & \ok\\
25 & \no & \no & \no & \no & \no & \no & \no & \no & \no & \no & \ok & \ok & \hm & \no & \no & \no\\
26 & \no & \no & \no & \no & \hm & \no & \hm & \no & \no & \no & \no & \no & \ok & \ok & \hm & \no\\
27 & \no & \no & \no & \no & \no & \no & \no & \no & \no & \no & \no & \no & \no & \no & \no & \no\\
28 & \no & \no & \no & \no & \no & \no & \no & \no & \ok & \ok & \ok & \ok & \ok & \ok & \ok & \ok\\
29 & \no & \no & \ok & \ok & \ok & \ok & \ok & \ok & \ok & \ok & \ok & \ok & \ok & \ok & \ok & \ok\\
30 & \no & \no & \no & \no & \no & \no & \no & \no & \no & \no & \ok & \ok & \no & \no & \hm & \no\\
31 & \no & \no & \no & \no & \no & \no & \no & \no & \ok & \ok & \ok & \ok & \ok & \ok & \ok & \ok\\
32 & \no & \no & \ok & \ok & \ok & \ok & \ok & \ok & \ok & \ok & \ok & \ok & \ok & \ok & \ok & \ok\\
33 & \hm & \no & \no & \no & \hm & \no & \no & \no & \ok & \ok & \ok & \ok & \ok & \ok & \ok & \ok\\
34 & \no & \no & \no & \no & \no & \no & \no & \no & \ok & \ok & \ok & \ok & \ok & \ok & \ok & \ok\\
35 & \no & \no & \ok & \ok & \ok & \ok & \ok & \ok & \ok & \ok & \ok & \ok & \ok & \ok & \ok & \ok\\
36 & \no & \no & \no & \no & \no & \no & \no & \no & \ok & \ok & \ok & \ok & \ok & \ok & \ok & \ok\\
37 & \no & \no & \no & \no & \no & \no & \no & \no & \ok & \ok & \ok & \ok & \ok & \ok & \ok & \ok\\
38 & \no & \no & \no & \no & \ok & \ok & \hm & \no & \ok & \ok & \ok & \ok & \ok & \ok & \ok & \ok\\
39 & \no & \no & \ok & \ok & \ok & \ok & \ok & \ok & \ok & \ok & \ok & \ok & \ok & \ok & \ok & \ok\\
40 & \hm & \no & \no & \no & \no & \no & \ok & \ok & \ok & \ok & \ok & \ok & \ok & \ok & \ok & \ok\\
41 & \no & \no & \no & \no & \ok & \ok & \no & \no & \ok & \ok & \ok & \ok & \ok & \ok & \no & \no\\
42 & \no & \no & \ok & \ok & \ok & \ok & \ok & \ok & \ok & \ok & \ok & \ok & \ok & \ok & \ok & \ok\\
43 & \no & \no & \ok & \ok & \ok & \ok & \ok & \ok & \ok & \ok & \ok & \ok & \ok & \ok & \ok & \ok\\
44 & \no & \no & \no & \no & \ok & \ok & \no & \no & \ok & \ok & \ok & \ok & \ok & \ok & \ok & \ok\\
45 & \no & \no & \ok & \ok & \ok & \ok & \no & \no & \ok & \ok & \ok & \ok & \ok & \ok & \ok & \ok\\
46 & \no & \no & \no & \no & \ok & \ok & \ok & \ok & \no & \hm & \ok & \ok & \ok & \ok & \ok & \ok\\
47 & \hm & \no & \no & \no & \no & \no & \no & \no & \no & \no & \ok & \ok & \ok & \ok & \no & \no\\
48 & \no & \no & \ok & \ok & \ok & \ok & \ok & \ok & \ok & \ok & \ok & \ok & \ok & \ok & \ok & \ok\\
49 & \no & \no & \ok & \ok & \ok & \ok & \ok & \ok & \ok & \ok & \ok & \ok & \ok & \ok & \ok & \ok\\
50 & \no & \no & \no & \no & \no & \no & \no & \no & \no & \no & \no & \no & \no & \no & \no & \no\\
\end{tabular}
\end{tiny}
\end{center}
\caption{Detailed results}
\label{fig:detailed_results}
\end{figure*}

\end{document}